# Continual Learning Through Synaptic Intelligence

Friedemann Zenke [* 1] Ben Poole [* 1] Surya Ganguli [1]

## Abstract

While deep learning has led to remarkable advances across diverse applications, it struggles in domains where the data distribution changes over the course of learning. In stark contrast, biological neural networks continually adapt to changing domains, possibly by leveraging complex molecular machinery to solve many tasks simultaneously. In this study, we introduce *intelligent synapses* that bring some of this biological complexity into artificial neural networks. Each synapse accumulates task relevant information over time, and exploits this information to rapidly store new memories without forgetting old ones. We evaluate our approach on continual learning of classification tasks, and show that it dramatically reduces forgetting while maintaining computational efficiency.

## 1. Introduction

Artificial neural networks (ANNs) have become an indispensable asset for applied machine learning, rivaling human performance in a variety of domain-specific tasks (LeCun et al., 2015). Although originally inspired by biology (Rosenblatt, 1958; Fukushima & Miyake, 1982), the underlying design principles and learning methods differ substantially from biological neural networks. For instance, parameters of ANNs are learned on a dataset in the training phase, and then frozen and used statically on new data in the deployment or recall phase. To accommodate changes in the data distribution, ANNs typically have to be retrained on the entire dataset to avoid overfitting and catastrophic forgetting (Choy et al., 2006; Goodfellow et al., 2013).

On the other hand, biological neural networks exhibit *continual learning* in which they acquire new knowledge over a lifetime. It is therefore difficult to draw a clear line between a learning and recall phase. Somehow, our brains have evolved to learn from non-stationary data and to update internal memories or beliefs on-the-fly. While it is unknown how this feat is accomplished in the brain, it seems possible that the unparalleled biological performance in continual learning could rely on specific features implemented by the underlying biological wetware that are not currently implemented in ANNs.

[*]Equal contribution [1]Stanford University. Correspondence to: Friedemann Zenke <fzenke@stanford.edu>, Ben Poole <poole@cs.stanford.edu>.

Perhaps one of the greatest gaps in the design of modern ANNs versus biological neural networks lies in the complexity of synapses. In ANNs, individual synapses (weights) are typically described by a single scalar quantity. On the other hand, individual biological synapses make use of complex molecular machinery that can affect plasticity at different spatial and temporal scales (Redondo & Morris, 2011). While this complexity has been surmised to serve memory consolidation (Fusi et al., 2005; Lahiri & Ganguli, 2013; Zenke et al., 2015; Ziegler et al., 2015; Benna & Fusi, 2016), few studies have illustrated how it benefits learning in ANNs.

Here we study the role of internal synaptic dynamics to enable ANNs to learn sequences of classification tasks. While simple, scalar one-dimensional synapses suffer from catastrophic forgetting, in which the network forgets previously learned tasks when trained on a novel task, this problem can be largely alleviated by synapses with a more complex three-dimensional state space. In our model, the synaptic state tracks the past and current parameter value, and maintains an online estimate of the synapse's "importance" toward solving problems encountered in the past. Our importance measure can be computed efficiently and locally at each synapse during training, and represents the local contribution of each synapse to the change in the global loss. When the task changes, we consolidate the important synapses by preventing them from changing in future tasks. Thus learning in future tasks is mediated primarily by synapses that were unimportant for past tasks, thereby avoiding catastrophic forgetting of these past tasks.

## 2. Prior work

The problem of alleviating catastrophic forgetting has been addressed in many previous studies. These studies can be broadly partitioned into (1) architectural, (2) functional, and (3) structural approaches.

**Architectural** approaches to catastrophic forgetting alter the architecture of the network to reduce interference between tasks without altering the objective function. The simplest form of architectural regularization is freezing certain weights in the network so that they stay exactly the same (Razavian et al., 2014). A slightly more relaxed approach reduces the learning rate for layers shared with the original task while fine-tuning to avoid dramatic changes in the parameters (Donahue et al., 2014; Yosinski et al., 2014). Approaches using different nonlinearities like ReLU, MaxOut, and local winner-take-all have been shown to improve performance on permuted MNIST and sentiment analysis tasks (Srivastava et al., 2013; Goodfellow et al., 2013). Moreover, injecting noise to sparsify gradients using dropout also improves performance (Goodfellow et al., 2013). Recent work from Rusu et al. (2016) proposed more dramatic architectural changes where the entire network for the previous task is copied and augmented with new features while solving a new task. This entirely prevents forgetting on earlier tasks, but causes the architectural complexity to grow with the number of tasks.

**Functional** approaches to catastrophic forgetting add a regularization term to the objective that penalizes changes in the input-output function of the neural network. In Li & Hoiem (2016), the predictions of the previous task's network and the current network are encouraged to be similar when applied to data from the new task by using a form of knowledge distillation (Hinton et al., 2014). Similarly, Jung et al. (2016) regularize the $\ell_2$ distance between the final hidden activations instead of the knowledge distillation penalty. Both of these approaches to regularization aim to preserve aspects of the input-output mapping for the old task by storing or computing additional activations using the old task's parameters. This makes the functional approach to catastrophic forgetting computationally expensive as it requires computing a forward pass through the old task's network for every new data point.

The third technique, **structural** regularization, involves penalties on the parameters that encourage them to stay close to the parameters for the old task. Recently, Kirkpatrick et al. (2017) proposed elastic weight consolidation (EWC), a quadratic penalty on the difference between the parameters for the new and the old task. They used a diagonal weighting proportional to the diagonal of the Fisher information metric over the old parameters on the old task. Exactly computing the diagonal of the Fisher requires summing over all possible output labels and thus has complexity linear in the number of outputs. This limits the application of this approach to low-dimensional output spaces.

## 3. Synaptic framework

To tackle the problem of continual learning in neural networks, we sought to build a simple structural regularizer that could be computed online and implemented locally at each synapse. Specifically, we aim to endow each individual synapse with a local measure of "importance" in solving tasks the network has been trained on in the past. When training on a new task we penalize changes to important parameters to avoid old memories from being overwritten. To that end, we developed a class of algorithms which keep track of an *importance measure* $\omega_k^\mu$ which reflects past credit for improvements of the task objective $L_\mu$ for task $\mu$ to individual synapses $\theta_k$. For brevity we use the term "synapse" synonymously with the term "parameter", which includes weights between layers as well as biases.

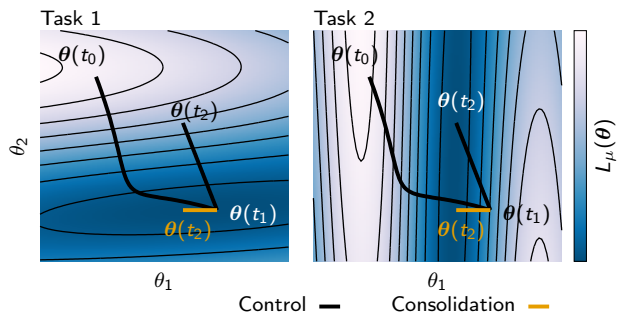

*Figure 1.* Schematic illustration of parameter space trajectories and catastrophic forgetting. Solid lines correspond to parameter trajectories during training. Left and right panels correspond to the different loss functions defined by different tasks (Task 1 and Task 2). The value of each loss function $L_\mu$ is shown as a heat map. Gradient descent learning on Task 1 induces a motion in parameter space from from $\boldsymbol{\theta}(t_0)$ to $\boldsymbol{\theta}(t_1)$. Subsequent gradient descent dynamics on Task 2 yields a motion in parameter space from $\boldsymbol{\theta}(t_1)$ to $\boldsymbol{\theta}(t_2)$. This final point minimizes the loss on Task 2 at the expense of significantly increasing the loss on Task 1, thereby leading to catastrophic forgetting of Task 1. However, there does exist an alternate point $\boldsymbol{\theta}(t_2)$, labelled in orange, that achieves a small loss for *both* tasks. In the following we show how to find this alternate point by determining that the component $\theta_2$ was more important for solving Task 1 than $\theta_1$ and then preventing $\theta_2$ from changing much while solving Task 2. This leads to an online approach to avoiding catastrophic forgetting by consolidating changes in parameters that were important for solving past tasks, while allowing only the unimportant parameters to learn to solve future tasks.

The process of training a neural network is characterized by a trajectory $\boldsymbol{\theta}(t)$ in parameter space (Fig. 1). The feat of successful training lies in finding learning trajectories for which the endpoint lies close to a minimum of the loss function $L$ on all tasks. Let us first consider the change in loss for an infinitesimal parameter update $\boldsymbol{\delta}(t)$ at time $t$.

In this case the change in loss is well approximated by the gradient $g = \frac{\partial L}{\partial \theta}$ and we can write

$$L(\boldsymbol{\theta}(t) + \boldsymbol{\delta}(t)) - L(\boldsymbol{\theta}(t)) \approx \sum_k g_k(t)\delta_k(t),  \quad (1)$$

which illustrates that each parameter change $\delta_k(t) = \theta'_k(t)$ contributes the amount $g_k(t)\delta_k(t)$ to the change in total loss.

To compute the change in loss over an entire trajectory through parameter space we have to sum over all infinitesimal changes. This amounts to computing the path integral of the gradient vector field along the parameter trajectory from the initial point (at time $t_0$) to the final point (at time $t_1$):

$$\int_C \boldsymbol{g}(\boldsymbol{\theta}(t))d\boldsymbol{\theta} = \int_{t_0}^{t_1} \boldsymbol{g}(\boldsymbol{\theta}(t)) \cdot \boldsymbol{\theta}'(t)dt. \quad (2)$$

As the gradient is a conservative field, the value of the integral is equal to the difference in loss between the end point and start point: $L(\boldsymbol{\theta}(t_1)) - L(\boldsymbol{\theta}(t_0))$. Crucial to our approach, we can decompose Eq. 2 as a sum over the individual parameters

$$\int_{t^{\mu-1}}^{t^\mu} \boldsymbol{g}(\boldsymbol{\theta}(t)) \cdot \boldsymbol{\theta}'(t)dt = \sum_k \int_{t^{\mu-1}}^{t^\mu} g_k(\theta(t))\theta'_k(t)dt$$
$$\equiv -\sum_k \omega_k^\mu. \quad (3)$$

The $\omega_k^\mu$ now have an intuitive interpretation as the parameter specific contribution to changes in the total loss. Note that we have introduced the minus sign in the second line, because we are typically interested in decreasing the loss.

In practice, we can approximate $\omega_k^\mu$ online as the running sum of the product of the gradient $g_k(t) = \frac{\partial L}{\partial \theta_k}$ with the parameter update $\theta'_k(t) = \frac{\partial \theta_k}{\partial t}$. For batch gradient descent with an infinitesimal learning rate, $\omega_k^\mu$ can be directly interpreted as the per-parameter contribution to changes in the total loss. In most cases the true gradient is approximated by stochastic gradient descent (SGD), resulting in an approximation that introduces noise into the estimate of $g_k$. As a direct consequence, the approximated per-parameter importances will typically overestimate the true value of $\omega_k^\mu$.

How can the knowledge of $\omega_k^\mu$ be exploited to improve continual learning? The problem we are trying to solve is to minimize the total loss function summed over all tasks, $\mathcal{L} = \sum_\mu L_\mu$, with the limitation that we do not have access to loss functions of tasks we were training on in the past. Instead, we only have access to the loss function $L_\mu$ for a single task $\mu$ at any given time. Catastrophic forgetting arises when minimizing $L_\mu$ inadvertently leads to substantial increases of the cost on previous tasks $L_\nu$ with $\nu < \mu$ (Fig. 1). To avoid catastrophic forgetting of all previous tasks ($\nu < \mu$) while training task $\mu$, we want to avoid drastic changes to weights which were particularly influential in the past. The importance of a parameter $\theta_k$ for a single task is determined by two quantities: 1) how much an individual parameter contributed to a drop in the loss $\omega_k^\nu$ over the entire trajectory of training (cf. Eq. 3) and 2) how far it moved $\Delta_k^\nu \equiv \theta_k(t^\nu) - \theta_k(t^{\nu-1})$. To avoid large changes to important parameters, we use a modified cost function $\tilde{L}_\mu$ in which we introduced a surrogate loss which approximates the summed loss functions of previous tasks $L_\nu$ ($\nu < \mu$). Specifically, we use a quadratic surrogate loss that has the same minimum as the cost function of the previous tasks and yields the same $\omega_k^\nu$ over the parameter distance $\Delta_k$. In other words, if learning were to be performed on the surrogate loss instead of the actual loss function, it would result in the same final parameters and change in loss during training (Fig. 2). For two tasks this is achieved exactly by the following quadratic surrogate loss

$$\tilde{L}_\mu = L_\mu + c \underbrace{\sum_k \Omega_k^\mu \left(\tilde{\theta}_k - \theta_k\right)^2}_{\text{surrogate loss}} \quad (4)$$

where we have introduced the dimensionless strength parameter $c$, the reference weight corresponding to the parameters at the end of the previous task $\tilde{\theta}_k = \theta_k(t^{\mu-1})$, and the per-parameter regularization strength:

$$\Omega_k^\mu = \sum_{\nu<\mu} \frac{\omega_k^\nu}{(\Delta_k^\nu)^2 + \xi} \quad . \tag{5}$$

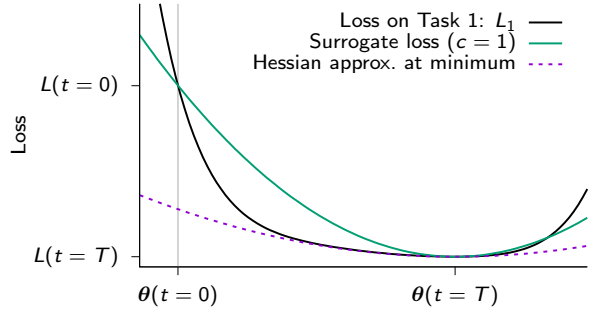

*Figure 2.* Schematic illustration of surrogate loss after learning one task. Consider some loss function defined by Task 1 (black). The quadratic surrogate loss (green) is chosen to precisely match 3 aspects of the descent dynamics on the original loss function: the total drop in the loss function $L(\boldsymbol{\theta}(0)) - L(\boldsymbol{\theta}(T))$, the total net motion in parameter space $\boldsymbol{\theta}(0) - \boldsymbol{\theta}(T)$, and achieving a minimum at the endpoint $\boldsymbol{\theta}(T)$. These 3 conditions uniquely determine the surrogate quadratic loss that summarizes the descent trajectory on the original loss. Note that this surrogate loss is different from a quadratic approximation defined by the Hessian at the minimum (purple dashed line).

Note that the term in the denominator $(\Delta_k^\nu)^2$ ensures that the regularization term carries the same units as the loss $L$. For practical reasons we also introduce an additional damping parameter, $\xi$, to bound the expression in cases where $\Delta_k^\nu \to 0$. Finally, $c$ is a strength parameter which trades off old versus new memories. If the path integral (Eq. 3) is evaluated precisely, $c=1$ would correspond to an equal weighting of old and new memories. However, due to noise in the evaluation of the path integral (Eq. 3), $c$ typically has to be chosen smaller than one to compensate. Unless otherwise stated, the $\omega_k$ are updated continuously during training, whereas the cumulative importance measures, $\Omega_k^\mu$, and the reference weights, $\tilde{\theta}$, are only updated at the end of each task. After updating the $\Omega_k^\mu$, the $\omega_k$ are set to zero. Although our motivation for Eq. 4 as a surrogate loss only holds in the case of two tasks, we will show empirically that our approach leads to good performance when learning additional tasks.

To understand how the particular choices of Eqs. 4 and 5 affect learning, let us consider the example illustrated in Figure 1 in which we learn two tasks. We first train on Task 1. At time $t_1$ the parameters have approached a local minimum of the Task 1 loss $L_1$. But, the same parameter configuration is not close to a minimum for Task 2. Consequently, when training on Task 2 without any additional precautions, the $L_1$ loss may inadvertently increase (Fig. 1, black trajectory). However, when $\theta_2$ "remembers" that it was important to decreasing $L_1$, it can exploit this knowledge during training on Task 2 by staying close to its current value (Fig. 1, orange trajectory). While this will almost inevitably result in a decreased performance on Task 2, this decrease could be negligible, whereas the gain in performance on both tasks combined can be substantial.

The approach presented here is similar to EWC (Kirkpatrick et al., 2017) in that more influential parameters are pulled back more strongly towards a reference weight with which good performance was achieved on previous tasks. However, in contrast to EWC, here we are putting forward a method which computes an importance measure online and along the *entire learning trajectory*, whereas EWC relies on a point estimate of the diagonal of the Fisher information metric at the final parameter values, which has to be computed during a separate phase at the end of each task.

## 4. Theoretical analysis of special cases

In the following we illustrate that our general approach recovers sensible $\Omega_k^\mu$ in the case of a simple and analytically tractable training scenario. To that end, we analyze what the parameter specific path integral $\omega_k^u$ and its normalized version $\Omega_k^\mu$ (Eq. (5)), correspond to in terms of the geometry of a simple quadratic error function

$$E(\boldsymbol{\theta}) = \frac{1}{2}(\boldsymbol{\theta} - \boldsymbol{\theta}^*)^T \boldsymbol{H}(\boldsymbol{\theta} - \boldsymbol{\theta}^*), \tag{6}$$

with a minimum at $\boldsymbol{\theta}^*$ and a Hessian matrix $\boldsymbol{H}$. Further consider batch gradient descent dynamics on this error function. In the limit of small discrete time learning rates, this descent dynamics is described by the continuous time differential equation

$$\tau \frac{d\boldsymbol{\theta}}{dt} = -\frac{\partial E}{\partial \boldsymbol{\theta}} = -\boldsymbol{H}(\boldsymbol{\theta} - \boldsymbol{\theta}^*), \tag{7}$$

where $\tau$ is related to the learning rate. If we start from an initial condition $\boldsymbol{\theta}(0)$ at time $t=0$, an exact solution to the descent path is given by

$$\boldsymbol{\theta}(t) = \boldsymbol{\theta}^* + e^{-\boldsymbol{H}\frac{t}{\tau}}(\boldsymbol{\theta}(0) - \boldsymbol{\theta}^*), \tag{8}$$

yielding the time dependent update direction

$$\boldsymbol{\theta}'(t) = \frac{d\boldsymbol{\theta}}{dt} = -\frac{1}{\tau}\boldsymbol{H}e^{-\boldsymbol{H}\frac{t}{\tau}}(\boldsymbol{\theta}(0) - \boldsymbol{\theta}^*). \tag{9}$$

Now, under gradient descent dynamics, the gradient obeys $\boldsymbol{g} = \tau \frac{d\boldsymbol{\theta}}{dt}$, so the $\omega_k^\mu$ in (3) are computed as the diagonal elements of the matrix

$$\boldsymbol{Q} = \tau \int_0^\infty dt \, \frac{d\boldsymbol{\theta}}{dt} \frac{d\boldsymbol{\theta}}{dt}^T . \tag{10}$$

An explicit formula for $\boldsymbol{Q}$ can be given in terms of the eigenbasis of the Hessian $\boldsymbol{H}$. In particular, let $\lambda^\alpha$ and $\boldsymbol{u}^\alpha$ denote the eigenvalues and eigenvectors of $\boldsymbol{H}$, and let $d^\alpha = \boldsymbol{u}^\alpha \cdot (\boldsymbol{\theta}(0) - \boldsymbol{\theta}^*)$ be the projection of the discrepancy between initial and final parameters onto the $\alpha$'th eigenvector. Then inserting (9) into (10), performing the change of basis to the eigenmodes of $\boldsymbol{H}$, and doing the integral yields

$$\boldsymbol{Q}_{ij} = \sum_{\alpha\beta} \boldsymbol{u}_i^\alpha d^\alpha \frac{\lambda^\alpha \lambda^\beta}{\lambda^\alpha + \lambda^\beta} d^\beta \boldsymbol{u}_j^\beta. \tag{11}$$

Note that as a time-integrated steady state quantity, $\boldsymbol{Q}$ no longer depends on the time constant $\tau$ governing the speed of the descent path.

At first glance, the $\boldsymbol{Q}$ matrix elements depend in a complex manner on both the eigenvectors and eigenvalues of the Hessian, as well as the initial condition $\boldsymbol{\theta}(0)$. To understand this dependence, let's first consider averaging $\boldsymbol{Q}$ over random initial conditions $\boldsymbol{\theta}(0)$, such that the collection of discrepancies $d^\alpha$ constitute a set of zero mean iid random variables with variance $\sigma^2$. Thus we have the average $\langle d^\alpha d^\beta \rangle = \sigma^2 \delta_{\alpha\beta}$. Performing this average over $\boldsymbol{Q}$ then yields

$$\langle \boldsymbol{Q}_{ij} \rangle = \frac{1}{2}\sigma^2 \sum_\alpha \boldsymbol{u}_i^\alpha \lambda^\alpha \boldsymbol{u}_j^\beta = \frac{1}{2}\sigma^2 \boldsymbol{H}_{ij}. \tag{12}$$

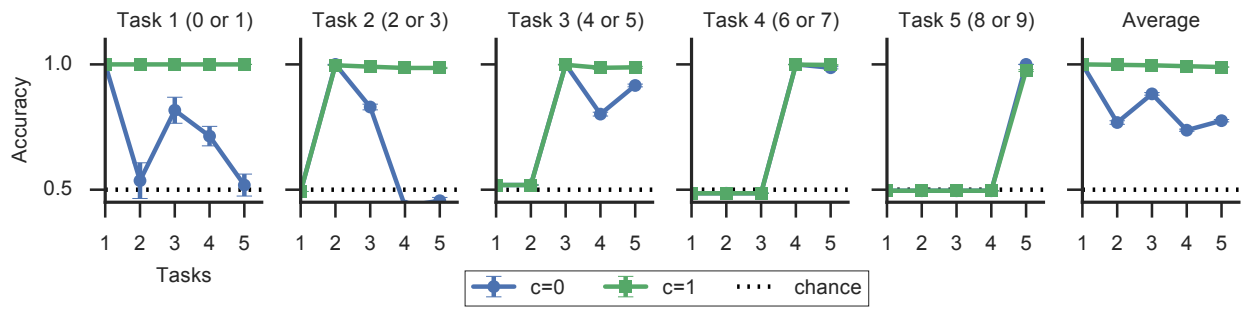

*Figure 3.* Mean classification accuracy for the split MNIST benchmark as a function of the number of tasks. The first five panels show classification accuracy on the five tasks consisting of two MNIST digits each as a function of number of consecutive tasks. The rightmost panel shows the average accuracy, which is computed as the average over task accuracies for past tasks $\nu$ with $\nu < \mu$ where $\mu$ is given by the number of tasks on the x-axis. Note that in this setup with multiple binary readout heads, an accuracy of 0.5 corresponds to chance level. Error bars correspond to SEM (n=10).

Thus remarkably, after averaging over initial conditions, the $Q$ matrix, which is available simply by correlating parameter updates across pairs of synapses and integrating over time, reduces to the Hessian, up to a scale factor dictating the discrepancy between initial and final conditions. Indeed, this scale factor theoretically motivates the normalization in (5); the denominator in (5), at zero damping, $\xi$ averages to $\sigma^2$, thereby removing the scale factor $\sigma^2$ in (12).

However, we are interested in what $Q_{ij}$ computes for a *single* initial condition. There are two scenarios in which the simple relationship between $Q$ and the Hessian $H$ is preserved *without* averaging over initial conditions. First, consider the case when the Hessian is diagonal, so that $u_i^\alpha = \delta_{\alpha i} e_i$ where $e_i$ is the $i$'th coordinate vector. Then $\alpha$ and $i$ indices are interchangeable and the eigenvalues of the Hessian are the diagonal elements of the Hessian: $\lambda^i = H_{ii}$. Then (11) reduces to

$$Q_{ij} = \delta_{ij}(d^i)^2 H_{ii}. \quad (13)$$

Again the normalization in (5), at zero damping, removes the scale of movement in parameter space $(d^i)^2$, and so the normalized $Q$ matrix becomes identical to the diagonal Hessian. In the second scenario, consider the extreme limit where the Hessian is rank 1 so that $\lambda^1$ is the only nonzero eigenvalue. Then (11) reduces to

$$Q_{ij} = \frac{1}{2}(d^1)^2 u_i^1 \lambda_1 u_j^1 = \frac{1}{2}(d^1)^2 H_{ij}. \quad (14)$$

Thus again, the $Q$ matrix reduces to the Hessian, up to a scale factor. The normalized importances then become the diagonal elements of the non-diagonal but low rank Hessian. We note that the low rank Hessian is the interesting case for continual learning; low rank structure in the error function leaves many directions in synaptic weight space unconstrained by a given task, leaving open excess capacity for synaptic modification to solve future tasks without interfering with performance on an old task.

It is important to stress that the path integral for importance is computed by integrating information along the *entire* learning trajectory (cf. Fig. 2). For a quadratic loss function, the Hessian is constant along this trajectory, and so we find a precise relationship between the importance and the Hessian. But for more general loss functions, where the Hessian varies along the trajectory, we cannot expect any simple mathematical correspondence between the importance $\Omega_k^\mu$ and the Hessian at the endpoint of learning, or related measures of parameter sensitivity (Pascanu & Bengio, 2013; Martens, 2016; Kirkpatrick et al., 2017) at the endpoint. In practice, however, we find that our importance measure is correlated to measures based on such endpoint estimates, which may explain their comparable effectiveness as we will see in the next section.

## 5. Experiments

We evaluated our approach for continual learning on the split and permuted MNIST (LeCun et al., 1998; Goodfellow et al., 2013), and split versions of CIFAR-10 and CIFAR-100 (Krizhevsky & Hinton, 2009).

### 5.1. Split MNIST

We first evaluated our algorithm on a split MNIST benchmark. For this benchmark we split the full MNIST training data set into 5 subsets of consecutive digits. The 5 tasks correspond to learning to distinguish between two consecutive digits from 0 to 10. We used a small multi-layer perceptron (MLP) with only two hidden layers consisting of 256 units each with ReLU nonlinearities, and a standard categorical cross-entropy loss function plus our consolidation cost term (with damping parameter $\xi = 1 \times 10^{-3}$). To avoid the complication of crosstalk between digits at the readout layer due to changes in the label distribution during training, we used a multi-head approach in which the categorical cross entropy loss at the readout layer was computed only for the digits present in the current task. Finally, we optimized our network using a minibatch size of 64 and trained for 10 epochs. To achieve good absolute performance with a smaller number of epochs we used the adaptive optimizer Adam (Kingma & Ba, 2014) ($\eta = 1 \times 10^{-3}$, $\beta_1 = 0.9$, $\beta_2 = 0.999$). In this benchmark the optimizer state was reset after training each task.

To evaluate the performance, we computed the average classification accuracy on all previous tasks as a function of number of tasks trained. We now compare this performance between networks in which we turn consolidation dynamics on ($c = 1$) against cases in which consolidation was off ($c = 0$). During training of the first task the consolidation penalty is zero for both cases because there is no past experience that synapses could be regularized against. When trained on the digits "2" and "3" (Task 2), both the model with and without consolidation show accuracies close to 1 on Task 2. However, on average the networks without synaptic consolidation show substantial loss in accuracy on Task 1 (Fig. 3). In contrast to that, networks with consolidation only undergo minor impairment with respect to accuracy on Task 1 and the average accuracy for both tasks stays close to 1. Similarly, when the network has seen all MNIST digits, on average, the accuracy on the first two tasks, corresponding to the first four digits, has dropped back to chance levels in the cases without consolidation whereas the model with consolidation only shows minor degradation in performance on these tasks (Fig. 3).

### 5.2. Permuted MNIST benchmark

In this benchmark, we randomly permute all MNIST pixels differently for each task. We trained a MLP with two hidden layers with 2000 ReLUs each and softmax loss. We used Adam with the same parameters as before. However, here we used $\xi = 0.1$ and the value for $c = 0.1$ was determined via a coarse grid search on a heldout validation set. The mini batch size was set to 256 and we trained for 20 epochs. In contrast to the split MNIST benchmark we obtained better results by maintaining the state of the Adam optimizer between tasks. The final test error was computed on data from the MNIST test set. Performance is measured by the ability of the network to solve all tasks.

To establish a baseline for comparison we first trained a network without synaptic consolidation ($c = 0$) on all tasks sequentially. In this scenario the system exhibits catastrophic forgetting, i.e. it learns to solve the most recent task, but rapidly forgets about previous tasks (blue line, Fig. 4). In contrast to that, when enabling synaptic consolidation, with a sensible choice for $c > 0$, the same network retains high classification accuracy on Task 1 while being trained on 9 additional tasks (Fig. 4). Moreover, the network learns to solve all other tasks with high accuracy and performs only slightly worse than a network which had trained on all data simultaneously (Fig. 4). Finally, these results were consistent across training and validation error and comparable to the results reported with EWC (Kirkpatrick et al., 2017).

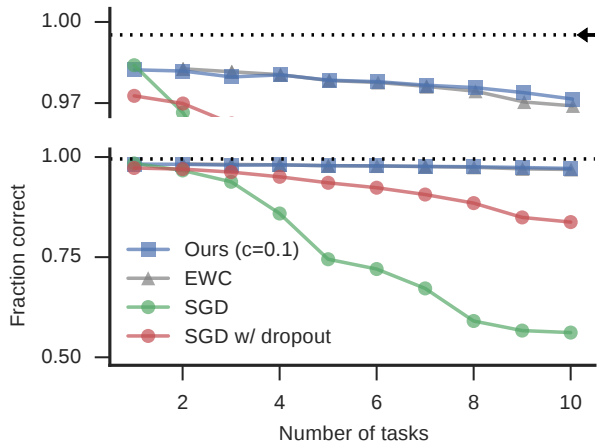

Figure 4. Average classification accuracy over all learned tasks from the permuted MNIST benchmark as a function of number of tasks. Our approach (blue) and EWC (gray, extracted and re-plotted from Kirkpatrick et al. (2017)) maintain high accuracy as the number of tasks increase. SGD (green) and SGD with dropout of 0.5 on the hidden layers (red) perform far worse. The top panel is a zoom-in on the upper part of the graph with the initial training accuracy on a single task (dotted line) and the training accuracy of the same network when trained on all tasks simultaneously (black arrow).

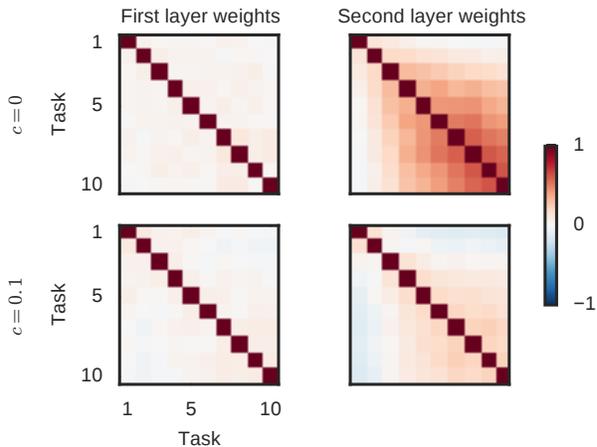

Figure 5. Correlation matrices of weight importances, $\omega_k^\mu$, for each task $\mu$ on permuted MNIST. For both normal fine-tuning ($c = 0$, top) and consolidation ($c = 0.1$, bottom), the first layer weight importances (left) are uncorrelated between tasks since the permuted MNIST datasets are uncorrelated at the input layer. However, the second layer importances (right) become more correlated as more tasks are learned with fine-tuning. In contrast, consolidation prevents strong correlations in the $\omega_k^\mu$, consistent with the notion of different weights being used to solve new tasks.

To gain a better understanding of the synaptic dynamics during training, we visualized the pairwise correlations of the $\omega_k^\mu$ across the different tasks $\mu$ (Fig. 5b). We found that without consolidation, the $\omega_k^\mu$ in the second hidden layer are correlated across tasks which is likely to be the cause of catastrophic forgetting. With consolidation, however, these sets of synapses contributing to decreasing the loss are largely uncorrelated across tasks, thus avoiding interference when updating weights to solve new tasks.

### 5.3. Split CIFAR-10/CIFAR-100 benchmark

To evaluate whether synaptic consolidation dynamics would also prevent catastrophic forgetting in more complex datasets and larger models, we experimented with a continual learning task based on CIFAR-10 and CIFAR-100. Specifically, we trained a CNN (4 convolutional, followed by 2 dense layers with dropout; see Appendix for details). We used the same multi-head setup as in the case of split MNIST using Adam ($\eta = 1 \times 10^{-3}$, $\beta_1 = 0.9$, $\beta_2 = 0.999$, minibatch size 256). First, we trained the network for 60 epochs on the full CIFAR-10 dataset (Task 1) and sequentially on 5 additional tasks each corresponding to 10 consecutive classes from the CIFAR-100 dataset (Fig. 6). To determine the best $c$, we performed this experiment for different values in the parameter range $1 \times 10^{-3} < c < 0.1$. Between tasks the state of the optimizer was reset. Moreover, we obtained values for two specific control cases. On the one hand we trained the same network with $c = 0$ on all tasks consecutively. On the other hand we trained the same network from scratch on each task individually to assess generalization across tasks. Finally, to assess the magnitude of statistical fluctuations in accuracy, all runs were repeated $n = 5$ times.

We found that after training on all tasks, networks with consolidation showed similar validation accuracy across all tasks, whereas accuracy in the network without consolidation showed a clear age dependent decline in which old tasks were solved with lower accuracy (Fig. 6). Importantly, the performance of networks trained with consolidation was always better than without consolidation, except on the last task. Finally, when comparing the performance of networks trained with consolidation on all tasks with net-works trained from scratch only on a single task (Fig. 6; green vs gray), the former either significantly outperformed the latter or yielded the same validation accuracy, while this trend was reversed in training accuracy. This suggests that networks without consolidation are more prone to over fitting. The only exception to that rule was Task 1, CIFAR-10 which is presumably due to its $10\times$ larger number of examples per class. In summary, we found that consolidation not only protected old memories from being slowly forgotten over time, but also allowed networks to generalize better on new tasks with limited data.

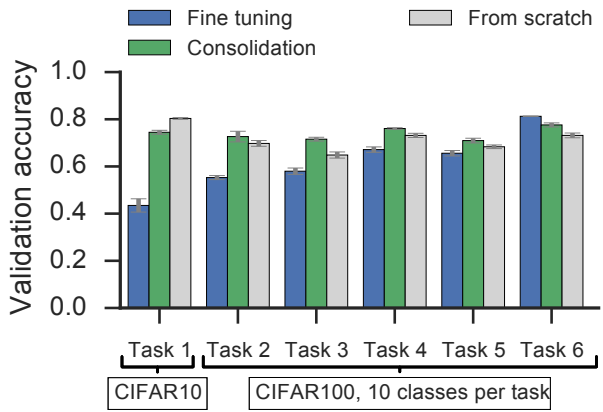

*Figure 6.* Validation accuracy on the split CIFAR-10/100 benchmark. Blue: Validation error, without consolidation ($c = 0$). Green: Validation error, with consolidation ($c = 0.1$). Gray: Network without consolidation trained from scratch on the single task only. Chance-level in this benchmark is 0.1. Error bars correspond to SD (n=5).

## 6. Discussion

We have shown that the problem of catastrophic forgetting commonly encountered in continual learning scenarios can be alleviated by allowing individual synapses to estimate their importance for solving past tasks. Then by penalizing changes to the most important synapses, novel tasks can be learned with minimal interference to previously learned tasks.

The regularization penalty is similar to EWC as recently introduced by Kirkpatrick et al. (2017). However, our approach computes the per-synapse consolidation strength in an online fashion and over the entire learning trajectory in parameter space, whereas for EWC synaptic importance is computed offline as the Fisher information at the minimum of the loss for a designated task. Despite this difference, these two approaches yielded similar performance on the permuted MNIST benchmark which may be due to correlations between the two different importance measures.

Our approach requires individual synapses to not simply correspond to single scalar synaptic weights, but rather act as higher dimensional dynamical systems in their own right. Such higher dimensional state enables each of our synapses to intelligently accumulate task relevant information during training and retain a memory of previous parameter values. While we make no claim that biological synapses behave like the intelligent synapses of our model, a wealth of experimental data in neurobiology suggests that biological synapses act in much more complex ways than the artificial scalar synapses that dominate current machine learning models. In essence, whether synaptic changes occur, and whether they are made permanent, or left to ultimately decay, can be controlled by many different biological factors. For instance, the induction of synaptic plasticity may depend on the history and the synaptic state of individual synapses (Montgomery & Madison, 2002). Moreover, recent synaptic changes may decay on the timescale of hours unless specific plasticity related chemical factors are released. These chemical factors are thought to encode the valence or novelty of a recent change (Redondo & Morris, 2011). Finally, recent synaptic changes can be reset by stereotypical neural activity, whereas older synaptic memories become increasingly insensitive to reversal (Zhou et al., 2003).

Here, we introduced one specific higher dimensional synaptic model to tackle a specific problem: catastrophic forgetting in continual learning. However, this suggests new directions of research in which we mirror neurobiology to endow individual synapses with potentially complex dynamical properties, that can be exploited to intelligently control learning dynamics in neural networks. In essence, in machine learning, in addition to adding depth to our networks, we may need to add intelligence to our synapses.

## Acknowledgements

The authors thank Subhaneil Lahiri for helpful discussions. FZ was supported by the SNSF (Swiss National Science Foundation) and the Wellcome Trust. BP was supported by a Stanford MBC IGERT Fellowship and Stanford Interdisciplinary Graduate Fellowship. SG was supported by the Burroughs Wellcome, McKnight, Simons and James S. McDonnell foundations and the Office of Naval Research.

## A. Split CIFAR-10/100 CNN architecture

For our CIFAR-10/100 experiments, we used the default CIFAR-10 CNN from Keras:

| Operation | Kernel | Stride | Filters | Dropout | Nonlin. |
|---|---|---|---|---|---|
| 3x32x32 input | | | | | |
| Convolution | $3 \times 3$ | $1 \times 1$ | 32 | | ReLU |
| Convolution | $3 \times 3$ | $1 \times 1$ | 32 | | ReLU |
| MaxPool | | $2 \times 2$ | | 0.25 | |
| Convolution | $3 \times 3$ | $1 \times 1$ | 64 | | ReLU |
| Convolution | $3 \times 3$ | $1 \times 1$ | 64 | | ReLU |
| MaxPool | | $2 \times 2$ | | 0.25 | |
| Dense | | | 512 | 0.5 | ReLU |
| Task 1: Dense | | | $m$ | | |
| ...: Dense | | | $m$ | | |
| Task $\mu$: Dense | | | $m$ | | |

Table 1. Split CIFAR10/100 model architecture and hyperparameters. $m$: number of splits.

## B. Additional split CIFAR–10 experiments

As an additional experiment, we trained a CNN (4 convolutional, followed by 2 dense layers with dropout; cf. main text) on the split CIFAR-10 benchmark. We used the same multi-head setup as in the case of split MNIST using Adam ($\eta = 1 \times 10^{-3}$, $\beta_1 = 0.9$, $\beta_2 = 0.999$, minibatch size 256). First, we trained the network for 60 epochs on the first 5 categories (Task A). At this point the training accuracy was close to 1. Then the optimizer was reset and the network was trained for another 60 epochs on the remaining 5 categories (Task B). We ran identical experiments for both the control case ($c = 0$) and the case in which consolidation was active ($c > 0$). All experiments were repeated $n = 10$ times to quantify the uncertainty on the validation set accuracy.

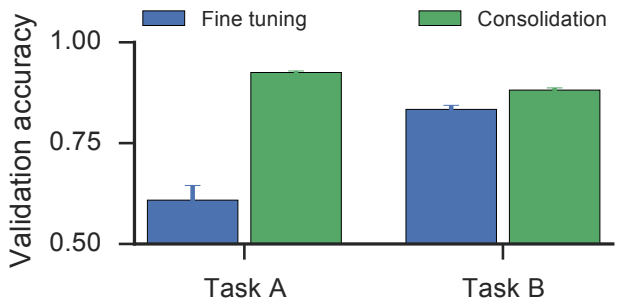

*Figure 7.* Classification accuracy for the split CIFAR-10 benchmark after being trained on Task A and B. Blue: Validation error, without consolidation ($c = 0$). Green: Validation error, with consolidation ($c = 0.1$). Note that chance-level in this benchmark is 0.2. Error bars correspond to SD (n=10).

After training on both Task A and B, the network with consolidation performed significantly better on both tasks than the control network without consolidation (Fig. 7). While the large performance difference on Task A can readily be explained by the fact that consolidation alleviates the problem of catastrophic forgetting — the initial motivation for our model — the small but significant difference ($\approx 4.5\%$) in validation accuracy on Task B suggests that consolidation also improves transfer learning. The network without consolidation is essentially fine-tuning a model which has been pre-trained on the first five CIFAR-10 categories. In contrast to that, by leveraging the knowledge about the optimization of Task A stored at the individual synapses, the network with consolidation solves a different optimization problem which makes the network generalize better on Task B. This significant effect was observed consistently for different values of $c$ in the range $0.1 < c < 10$.

## C. Comparison of path integral approach to other metrics

Prior approaches toward measuring the sensitivity of parameters in a network have primarily focused on local metrics related to the curvature of the objective function at the final parameters (Martens, 2016). The Hessian is one possible metric, but it can be negative definite and computing even the diagonal adds additional overhead over standard backpropagation (Martens et al., 2012). An alternative choice is the Fisher information (see for instance Kirkpatrick et al. (2017)):

$$\mathcal{F} = \mathbb{E}_{x \sim \mathcal{D}, y \sim p_\theta(y|x)} \left[ \left( \frac{\partial \log p_\theta(y|x)}{\partial \theta} \right) \left( \frac{\partial \log p_\theta(y|x)}{\partial \theta} \right)^T \right]$$

While the Fisher information has a number of desirable properties (Pascanu & Bengio, 2013), it requires computing gradients using labels sampled from the model distribution instead of the data distribution, and thus would require at least one additional backpropagation pass to compute online. For efficiency, the Fisher is often replaced with an approximation, the empirical Fisher (Martens, 2016), that uses labels sampled from the data distribution and can be computed directly from the gradient of the objective at the current parameters:

$$\bar{\mathcal{F}} = \mathbb{E}_{(x,y) \sim \mathcal{D}} \left[ \left( \frac{\partial \log p_\theta(y|x)}{\partial \theta} \right) \left( \frac{\partial \log p_\theta(y|x)}{\partial \theta} \right)^T \right]$$
$$= \mathbb{E}_{(x,y) \sim \mathcal{D}} \left[ g(\theta) g(\theta)^T \right]$$

The diagonal of the empirical Fisher yields a very similar formula to our local importance measure $\omega$ in Eq. 3 under gradient descent dynamics. However, the empirical Fisher is computed at a single parameter value $\boldsymbol{\theta}$ whereas the path integral is computed over a trajectory $\boldsymbol{\theta}(t)$. This yields an important difference in the behavior of these metrics: for a quadratic the empirical Fisher at the minimum will be 0 while the path integral will be proportional to the diagonal of the Hessian. Thus the path integral based approach yields an efficient algorithm with no additional gradients required that still recovers a meaningful estimate of the curvature.